\pdfoutput=1

\documentclass[11pt]{article}

\usepackage[]{ACL2023}

\usepackage{times}
\usepackage{latexsym}

\usepackage[T1]{fontenc}

\usepackage[utf8]{inputenc}

\usepackage{microtype}

\usepackage{inconsolata}

\usepackage{adjustbox}
\usepackage{graphicx}
\usepackage{xspace}
\usepackage{makecell}
\usepackage{booktabs}
\usepackage{multirow}

\newcommand{\kpk}{\texttt{\textsc{KP20k}}\xspace}
\newcommand{\openkp}{\texttt{\textsc{OpenKP}}\xspace}
\newcommand{\kptimes}{\texttt{\textsc{KPTimes}}\xspace}
\newcommand{\stackex}{\texttt{\textsc{StackEx}}\xspace}
\newcommand{\jptimes}{\texttt{\textsc{JPTimes}}\xspace}
\newcommand{\duc}{\texttt{\textsc{DUC-2001}}\xspace}
\newcommand{\magcs}{\texttt{\textsc{MAG-CS}}\xspace}
\newcommand{\onetoseq}{\texttt{One2Seq}\xspace}

\newcommand{\bos}{\texttt{<bos>}\xspace}
\newcommand{\sep}{\texttt{<sep>}\xspace}
\newcommand{\eos}{\texttt{<eos>}\xspace}

\newcommand{\pt}{\textbf{PT}\xspace}
\newcommand{\da}{\textbf{DA}\xspace}
\newcommand{\ft}{\textbf{FT}\xspace}

\newcommand{\tfrand}{\texttt{\textsc{TF-Rand}}\xspace}
\newcommand{\tfbart}{\texttt{\textsc{TF-Bart}}\xspace}

\usepackage{array}
\newcommand{\PreserveBackslash}[1]{\let\temp=\\#1\let\\=\temp}
\newcolumntype{C}[1]{>{\PreserveBackslash\centering}p{#1}}
\newcolumntype{R}[1]{>{\PreserveBackslash\raggedleft}p{#1}}
\newcolumntype{L}[1]{>{\PreserveBackslash\raggedright}p{#1}}

\title{General-to-Specific Transfer Labeling for \\Domain Adaptable Keyphrase Generation}

\author{
    Rui Meng\textsuperscript{\rm 1},
    Tong Wang\textsuperscript{\rm 2},
    Xingdi Yuan\textsuperscript{\rm 2},
    Yingbo Zhou\textsuperscript{\rm 1},
    Daqing He\textsuperscript{\rm 3}
    \\
    \textsuperscript{\rm 1}Salesforce Research, \textsuperscript{\rm 2}Microsoft Research, Montr\'{e}al, \textsuperscript{\rm 3}University of Pittsburgh
    \\
    \texttt{ruimeng@salesforce.com}
    }

\begin{document}
\maketitle
\begin{abstract}

Training keyphrase generation (KPG) models require a large amount of annotated data, which can be prohibitively expensive and often limited to specific domains.
In this study, we first demonstrate that large distribution shifts among different domains severely hinder the transferability of KPG models.
We then propose a three-stage pipeline, which gradually guides KPG models' learning focus from general syntactical features to domain-related semantics, in a data-efficient manner. 
With domain-general phrase pre-training, we pre-train Sequence-to-Sequence models with generic phrase annotations that are widely available on the web, which enables the models to generate phrases in a wide range of domains.
The resulting model is then applied in the Transfer Labeling stage to produce domain-specific pseudo keyphrases, which help adapt models to a new domain.
Finally, we fine-tune the model with limited data with true labels to fully adapt it to the target domain.
Our experiment results show that the proposed process can produce good quality keyphrases in new domains and achieve consistent improvements after adaptation with limited in-domain annotated data\footnote{All code and datasets are available at 
\url{https://github.com/memray/OpenNMT-kpg-release}.}\footnote{The research was mostly accomplished when the first author was at the University of Pittsburgh.}.

\end{abstract}

\section{Introduction}

The last decade has seen major advances in deep neural networks and their applications in natural language processing. 
Particularly, the sub-area of neural keyphrase generation (KPG) has made great progress with the aid of large language models~\cite{lewis2020bart} and large-scale datasets~\cite{meng2017deep,yuan2020one}. 
Due to the high cost of data annotation, most, if not all, of the large-scale KPG datasets are constructed by scraping domain-specific data from the internet.
For example, \citeauthor{meng2017deep} collected more than 500k scientific papers of which keyphrases are provided by paper authors. 
\citeauthor{gallina2019kptimes} crawled about 280k news articles from New York Times with editor-assigned keyphrases.
Following \citet{gururangan2020don}, we use ``domain'' to denote a distribution over language characterizing a given topic or genre.
Specifically in KPG tasks, domains can be ``computer science papers'', ``online forum articles'', ``news'' etc.

Despite recent neural models can to some extent learn KPG skills from existing datasets \cite{meng-etal-2021-empirical, gallina2019kptimes, yuan2020one}, because most of these datasets are limited to a single domain, it remains unclear how the trained models can be transferred to new domains, especially in a real-world setting. 
Some existing studies claim their models demonstrate a certain degree of transferability across domains. 
For instance, \citeauthor{meng2017deep} show that models trained with scientific paper datasets can generate decent quality keyphrases from news articles, in a zero-shot manner. 
\citeauthor{xiong2019open} present that training with open-domain web documents can improve the model's generalizability.
However, there is a lack of systematic studies on domain transferring KPG, and thus the observations reported in prior works do not support a comprehensive understanding of this topic.

To investigate this question, we conduct an empirical study on how well KPG models can transfer across domains. 
We utilize commonly used KPG datasets covering four different domains (Science, News, Web, Q\&A).
We first show experiment results (\S\ref{sec:analyze-domain-gap}) that suggest models trained with data in a specific domain do not generalize well to other domains, even in cases where they are initialized with pre-trained language models such as BART~\cite{lewis2020bart}. 
We also visualize the domain gaps among datasets by inspecting their phrase overlaps.
Keyphrases often represent the specific knowledge of a domain and this may result in the failure of transferring models across domains.

The empirical study motivates us to explore novel methods that can help models possess the ability of generating high quality keyphrases and more importantly, can quickly adapt to a new domain with limited amount of annotation. 
We propose a three-stage training pipeline, in which we gradually guide a KPG model's learning focus from general syntactical features to domain-specific information. 
First, we pre-train the model using community labeled phrases in Wikipedia (\S\ref{sec:stage1}).
Then, we use a novel self-training-based domain adaptation method, namely Transfer Labeling, to adapt the model to the new domain.
Note this domain adaptation method does not require ground-truth labels, we leverage the model pre-trained in the previous stage to generate pseudo-labels for training itself. 
Finally, we use a limited amount of in-domain data with true annotations to fully adapt the model to the new domain.
We report extensive experiment results and thorough analyses to demonstrate the effectiveness of the proposed methods.

\section{Background and Motivation}
\label{sec:motivatons}

\subsection{Background}
\label{sec:background}

\paragraph{Keyphrase Generation (KPG)}
Typically, the task is to generate a set of keyphrases $P = \{p_1,\dots,p_n\}$ given a source text $t$.
Semantically, these phrases summarize and highlight important information contained in $t$,
while syntactically, each keyphrase may consist of multiple words and serve a component of a sentence.
Depending on a particular domain the source text belongs to (e.g., scientific paper, news) and downstream applications (e.g., article classification, information retrieval), the extent to which a phrase is important can vary, i.e. the criteria of keyphrase can be different in various datasets.
Following \citeauthor{meng2017deep}, we denote a keyphrase as \emph{present} if it is a sub-string of the source text, or as \emph{absent} otherwise.
We adopt the \onetoseq training paradigm \cite{yuan2020one}. 
Given a source text $t$ and a set of ground-truth keyphrases $P$, we concatenate all ground-truth keyphrases into a single string: $\bos p_1 \sep \cdots \sep p_n \eos$, where \bos, \sep, and \eos are special tokens. 
This string is paired with $t$ to train a sequence-to-sequence model.
We refer readers to \cite{meng-etal-2021-empirical} for more details in common KPG practice.

\subsection{Domain Gap in KPG Tasks}
\label{sec:analyze-domain-gap}
Previous studies have touched on how much KPG models can transfer their skills when applied across domains~\cite{meng2017deep,xiong2019open}, but not in a systematic way. 
In this subsection, we revisit this topic and try to ground our discussion with thorough empirical results. 
Specifically, we consider four broadly used datasets in the KPG community: \kpk~\cite{meng2017deep} contains scientific papers in computer science;
\openkp~\cite{xiong2019open} is a collection of web documents;
\kptimes~\cite{gallina2019kptimes} contains a set of news articles;
\stackex~\cite{yuan2020one} are community-based Q\&A posts 
collected from StackExchange.
All the four datasets are large enough to train KPG models from scratch.
At the same time, the documents in these datasets cover a wide spectrum of domains. 
We report statistics of these four datasets in appendix Table~\ref{tab:appendix-dataset-statistics}.


\begin{figure}[ht!]
    \centering
    \includegraphics[width=0.45\textwidth]{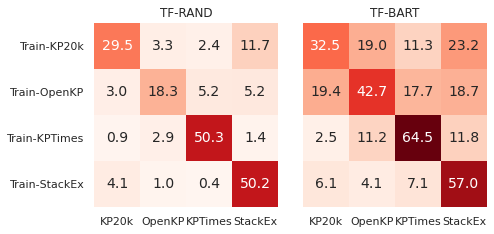}
    \caption{Cross-domain transfer performance of \tfrand and \tfbart (F@O, the higher the better). Y-axis: training dataset; X-axis: test dataset.}
    \label{fig:domain_transfer}
\end{figure}

On the model dimension, we consider two model architectures: \tfrand, a 6-layer encoder-decoder Transformer with random initialization~\cite{vaswani2017attention}; and \tfbart, a 12-layer Transformer initialized with BART-large~\cite{lewis2020bart}. 
We train the two models on the four datasets individually and subsequently evaluate all the resulting eight checkpoints on the test split of each dataset.
As shown in Figure~\ref{fig:domain_transfer}, in-domain scores (i.e., trained and tested on the same datasets) are placed along the diagonal, the other elements represent cross-domain testing scores. 
We observe that both models exhibit a large gap between in-domain and out-of-domain performance. 
Even though the initialization with BART can alleviate the gap to a certain degree, the difference remains significant.



\begin{table}[!htp]
    \vspace{-0.5em}
    \renewcommand{\arraystretch}{0.7}
    \setlength{\tabcolsep}{0.3em}
    \small
    \centering
    \begin{tabular}{l|c|c|c|c}
    \toprule
        ~ & \kpk & \openkp & \kptimes & \stackex \\ \midrule
        \kpk   & 100 & 10.6 & 3.5 & \textcolor{red}{55.7} \\ 
        \openkp  & 3.2 & 100  & 8.5 & 33.1 \\ 
        \kptimes & 0.5 & 4.3  & 100 & 6.9 \\ 
        \stackex & \textcolor{red}{0.7} & 1.3  & 0.5 & 100 \\
     \bottomrule
    \end{tabular}
    \vspace{-0.5em}
    \caption{Overlap (\%) of unique keyphrases between domains (train split). Numbers are normalized by dividing the diagonal element in its column. For example, the overlap keyphrases between \kpk and \stackex make a proportion of \textcolor{red}{0.7\%} in \kpk and a \textcolor{red}{55.7\%} in \stackex.}
    \vspace{-0.5em}
    \label{tab:phrase-vocab-overlap}
\end{table}


\begin{figure*}[!ht]
    \centering
    \includegraphics[width=0.7\textwidth]{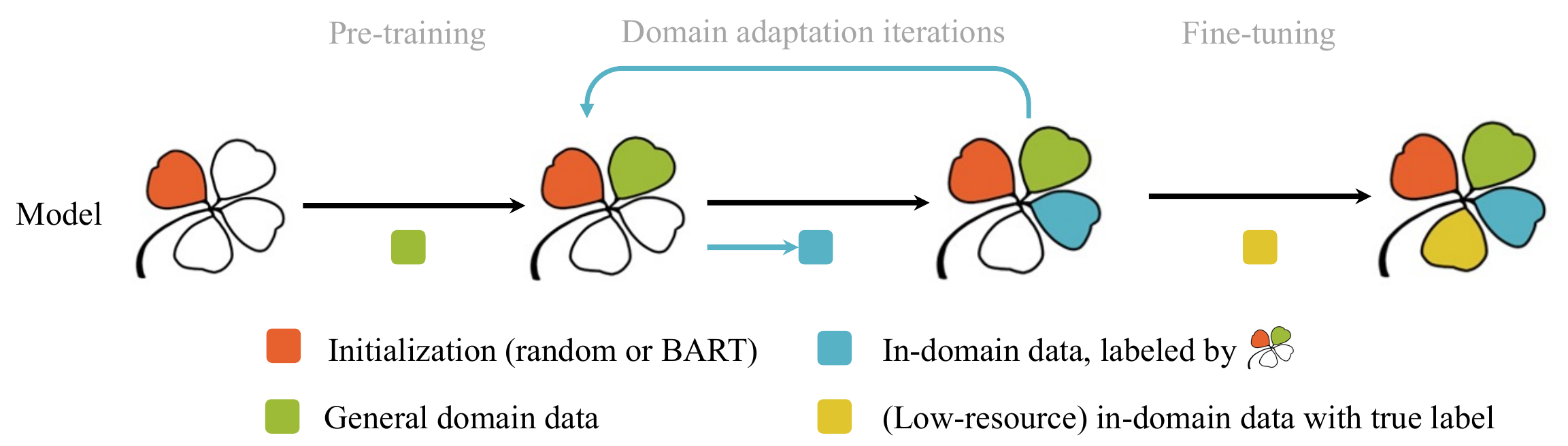}
    \caption{The proposed three-stage pipeline. 
    A model is first pre-trained with general domain data and learns to generate syntactically correct phrases.
    In the domain adaptation stage, the model adapts to the target domain by training on domain-specific data, where the pseudo labels are generated by the model itself.
    Finally, we fine-tune the model with limited amount of target domain data with true label, to fully accomplish domain adaptation.}
    \label{fig:pipeline}
\end{figure*}

Keyphrases are typically concepts or entities that represent important information of a document. 
The collection of keyphrases in a domain can also be deemed as a representation of domain knowledge. 
Therefore, to better investigate the domain gaps, we further look into the keyphrase overlap between datasets.
As shown in Table~\ref{tab:phrase-vocab-overlap}, only a small proportion of phrases are in common between the four domains.
We provide a T-SNE visualization of a set of phrases sampled from these dataset in appendix Figure~\ref{fig:appendix-visualize-domain-difference}, the phrase clusters present clear domain gaps in their semantic space.

We hypothesize that the domain specific traits in annotated data make models difficult to learn keyphrase patterns in a domain-general sense. 
Furthermore, humans may label keyphrases under an application-oriented consideration and thus a one-size-fits-all standard for keyphrase annotation may not exist. 
For example, on StackExchange, users tend to assign common tags to better expose their questions to community experts, resulting in a small keyphrase vocabulary size. 
On the contrary, the topics are more specialized in scientific papers and authors would emphasize novel concepts in their studies. This may explain the large number of unique keyphrases found in \kpk.

\subsection{Disentanglement of ``Key'' and ``Phrase''}
In \S\ref{sec:analyze-domain-gap}, we empirically show that KPG models do not adequately transfer to out-of-domain data, even initialized with pre-trained language models.
However, data annotation for every single domain or application does not seem practical either, due to the high cost and the potential need of domain-specific annotators. 
Inspired by some prior works, we attempt to disentangle the important properties of a keyphrase as \textit{keyness}~\cite{bondi2010keyness,gabrielatos2018keyness} and \textit{phraseness}~\cite{tomokiyo2003language}.
We believe a proficient KPG model should generate outputs that satisfy both properties.


\textbf{Keyness} refers to the attribute that how well a phrase represents important information of a piece of text. 
The degree of keyness can be document dependent and domain dependent. 
For example, ``cloud'' is a common keyphrase in Computer Science papers, it is, in most cases, less likely to be important in Meteorology studies. 
Due to its high dependence on domain-specific information, we believe that the knowledge/notion of keyness is more likely to be acquired from in-domain data.


\textbf{Phraseness}, on the other hand, focuses more on the syntactical aspect. 
It denotes that given a short piece of text, without even taking into account its context, to what extent it can be grammatically functional as a meaningful unit. 
Although the majority of keyphrases in existing datasets are noun phrases~\cite{chuang2012without}, they can present in variant grammatical forms in the real world~\cite{sun2021global}. 
We believe that phraseness can be independent from domains and thus can be obtained from domain-general data.

\section{Methodology}
\label{sec:methodology}

In the spirit of the motivation discussed above, we propose a three-stage training procedure in which a model gradually moves its focus from learning domain-general phraseness towards domain-specific keyness, and eventually adapts to a new domain with only limited amount of annotated data.
An overview of the proposed pipeline is illustrated in Figure~\ref{fig:pipeline}. 
First, with a Pre-Training stage (\pt), the model is trained with domain-general data to learn phraseness (\S\ref{sec:stage1}).
Subsequently, in the Domain Adaption stage (\da), the model is exposed with \textit{unlabeled} in-domain data.
Within a few iterations, the model labels the data itself and use them to gradually adapt to the new domain (\S\ref{sec:stage2}). 
Lastly, in the Fine-Tuning stage (\ft), the model fully adapts itself to the new domain by leveraging a limited amount of in-domain data with true annotations (\S\ref{sec:stage3}). 
In this section, we describe each of the three stages in detail.


\subsection{Domain-General Phrase Pre-training}
\label{sec:stage1}

The first training stage aims to capture the phraseness in general, we leverage the Wikipedia data and community labeled phrases from the text.
Wikipedia is an open-domain knowledge base that contains rich entity-centric annotations, its articles cover a wide spectrum of topics and domains and thus it has been extensively used as a resource of distant supervision for NLP tasks related to entities and knowledge~\cite{ghaddar-langlais-2017-winer,yamada2020luke, xiong2019pretrained}.
In this work, we consider four types of markup patterns in Wikipedia text to form distant keyphrase labels:

\begin{itemize}
  \setlength\itemsep{-0.2em}
  \item in-text phrases with special formatting (italic, boldface, and quotation marks); 
  \item in-text phrases with wikilinks (denoting an entity in Wikipedia); 
  \item ``see also'' phrases (denoting related entities); 
  \item ``categories'' phrases (denoting superordinate entities).
\end{itemize}
Although the constructed targets using the above heuristics can be noisy if considering the keyness aspect, we show that they work sufficiently for training general phrase generation models.



\begin{figure}[t!]
\centering
\includegraphics[width=0.35\textwidth]{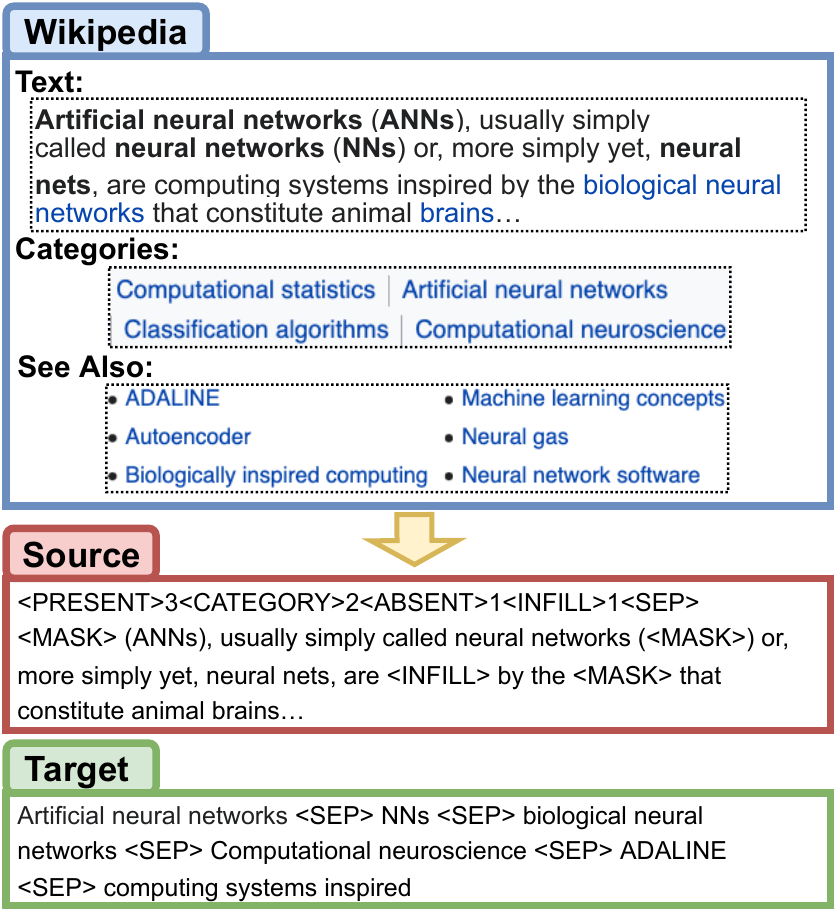}
\caption{Illustration of processing Wikipedia to source-target pairs in domain general phrase pre-training.}
\vspace{-1em}
\label{fig:wiki-process}
\end{figure}

Given a piece of Wikipedia text $t$ and a set of community labeled phrases, we convert this data point to the format of \onetoseq as described in \S\ref{sec:background}.
In practice, the number of phrases within $t$ can be large and thus we sample a subset from them to form the target. 
We group all the phrases appear in $t$ as present candidates, the rest (e.g., ``see-also'' and categories) are grouped as absent candidates.
Additionally, we take several random spans from $t$ as infilling candidates (similar as \cite{raffel2020t5}) for robustness. 
Finally, we sample a few candidates from each group and concatenate them as the final target sequence. 

On the source side, we prepend a string suggesting the cardinality of phrases in each target group to the beginning of $t$. 
We also corrupt the source sequence by replacing a small proportion of present and infilling phrases with a special token \texttt{[MASK]}, expecting to improve models' robustness \cite{raffel2020t5}.
We show an example of a processed Wikipedia data instance in Figure~\ref{fig:wiki-process}.


Trained with this data, we expect a model to become a general phrase generator --- given a source text, the model can generate a sequence of phrases, regardless the specific domain a text belongs to.

\subsection{Domain Adaption with Transfer Labeling}
\label{sec:stage2}
In the second stage, we aim to expose the model with data from a domain of interest, so it can learn the notion of domain-specific keyness.
We propose a method, namely General-to-Specific Transfer Labeling 
, which does not require any in-domain annotated data.
Transfer labeling can be considered as a special self-training method~\cite{yarowsky1995unsupervised,culp2008iterative,mukherjee2020uncertainty}, where the key notion is to train a model with its own predictions iteratively.

Distinct from common practice of self-training where initial models are bootstrapped with annotated data, transfer labeling regards the  domain-general model from the pre-training stage~\ref{sec:stage1} as a qualified phrase predictor.
We directly transfer the model to documents in a new domain to predict pseudo labels.
The resulting phrases, paired with these documents, are used to tune the model so as to adapt it to the target domain distribution.
Note that this process can be run iteratively, to gradually adapt models to target domains.


\subsection{Low-resource Fine-Tuning}
\label{sec:stage3}
In the third stage, we expose the model to a small amount of in-domain data with annotated keyphrases.
This aims to help the model fully adapt to the new domain and reduce the bias caused by noisy labels from previous stages.

\section{Experiments}
\label{sec:exp}

We reuse the model architecture described in \S\ref{sec:analyze-domain-gap} throughout this paper. And most models apply a single iteration of transfer labeling. We discuss the effect of multi-iteration transfer labeling in \S\ref{sec:multi-iter-da}. 
See Appendix~\ref{app:implementation} for implementation details.

\subsection{Datasets and Evaluation Metric}
\label{sec:data_and_metric}

We consider the same four large-scale KPG datasets as described in \S\ref{sec:analyze-domain-gap}, but instead of training models with all annotated document-keyphrases pairs, we take a large set of unannotated documents from each dataset for domain adaptation, and a small set of annotated examples for few-shot fine-tuning. 
Specifically, in the pre-training stage (\pt), we use the 2021-05-21 release of English Wikipedia dump and process it with wikiextractor package, which results in 3,247,850 passages.
In the domain adaptation stage (\da), for each domain, we take the first 100k examples from the training split (without keyphrases), and apply different strategies to produce pseudo labels and subsequently train the models.
In the fine-tuning stage (\ft), we take the first 100/1k/10k annotated examples (document-keyphrases pairs) from the training split to train the models. 
We report the statistics of used datasets in appendix Table~\ref{tab:appendix-dataset-statistics}.

We follow previous studies to split training/validation/test sets, and report model performance on test splits of each dataset.
A common practice in KPG studies is to evaluate the model performance on present/absent keyphrases separately. 
However, the ratios of present/absent keyphrases differ drastically among the four datasets (e.g. \openkp is strongly extraction-oriented).
Since we aim to improve the model's out-of-domain performance in general regardless of the keyphrases being present or absent, we follow \citet{bahuleyan2020unlikelihood} and simply evaluate present and absent keyphrases altogether. 
We report the F@O scores~\cite{yuan2020one} between the generated keyphrases and the ground-truth.
This metric requires systems to model the cardinality of predicted keyphrases themselves.


\subsection{Results and Analyses}

\subsubsection{Zero-shot Performance}
\label{sec:zeroshot}
We first investigate how well models can perform after the pre-training stage, without utilizing any in-domain annotated data. 
Since Wikipedia articles contain a rather wide range of phrase types, we expect models trained on this data are capable of predicting relevant and well-formed phrases from documents in general. 
We show our models' testing scores in the first row of Table~\ref{tab:fewshot-result-transformer} and \ref{tab:fewshot-result-bart-simple}, where only \pt is checked. 
We observe that pre-training with Wikipedia data can provide decent zero-shot performance in both settings, i.e.,  model is initialized randomly (Table~\ref{tab:fewshot-result-transformer}) and with pre-trained language models (\ref{tab:fewshot-result-bart-simple}).
Both settings achieve the same average F@O score of 12.2, which evinces the feasibility of using \pt model to generate pseudo labels for further domain adaptation. 
The scores also suggest that at the pre-training stage, the BART model (with pre-trained initialization and more parameters) does not present an advantage in comparison to a smaller model trained from scratch.


\subsubsection{Domain Adaptation Strategies}
\label{sec:DA_strategies}

\begin{figure}[t!]
    \centering
    \includegraphics[width=0.4\textwidth]{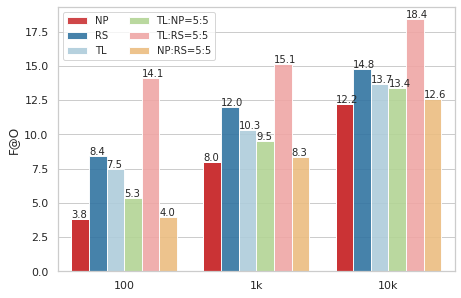}
    \vspace{-0.5em}
    \caption{Comparison of different strategies for domain adaptation with \tfrand.  \textbf{TL}: Transfer Labeling. \textbf{NP}: Noun Phrases. \textbf{RS}: Random Span.}
    \vspace{-1.0em}
    \label{fig:DA_strategies}
\end{figure}

We compare transfer labeling (\textbf{TL}, proposed in \S\ref{sec:stage2}) with two unsupervised strategies: (1) Noun Phrase (\textbf{NP}) and (2) Random Span (\textbf{RS}). 
For NP, we employ SpaCy~\cite{Honnibal2020spacy} to POS-tag source texts and extract noun phrases based on regular expressions. 
For RS, we follow \citet{raffel2020t5}, extracting random spans as targets and masking them in the source text. 
For TL, all pseudo phrases are generated by a \pt model in a zero-shot manner (with greedy decoding). 

As shown in Figure~\ref{fig:DA_strategies}, in the single strategy setting, RS performs the best among the three strategies and TL follows. 
We speculate that RS models are trained to predict randomly masked spans based on their context, and this results in the best generalization among the three. 
As for the NP strategy, since targets are only noun phrases appear in the source text, the models may have the risk of overfitting to recognize a subset of possible phrases.
TL lies in between the two discussed strategies, the generated pseudo labels contain both present and absent phrases, and thanks to the \pt model trained with Wikipedia data, the generated targets can contain many phrase types beyond noun phrases. 

We further investigate the performance gap between RS and TL. On \kpk, the \pt model can generate 5.1 present and 2.6 absent keyphrases on average.
The generated pseudo labels, albeit of good quality, are always fixed during the training. This is due to the deterministic nature of the \pt model, which may cause overfitting and limit the model's generalizability.
In contrast, random spans in RS are dynamically generated, therefore a model can learn to generate different target phrases even the same documents appear multiple times during training. 
This motivates us to investigate if these strategies can be synergistic by combining them. 
As shown in Figure~\ref{fig:DA_strategies}, we observe that combining TL and RS can lead to a significant improvement over all other strategies, indicating that these two strategies are somewhat complementary and thus can be used together in domain adaption. 
In the rest of the paper, we by default combine TL and RS in the domain adaptation stage, by taking equal amount of data from both sides, we discuss other mixing strategies in Appendix~\ref{app:more_results}.


It is worth noting that, if we apply domain adaptation with the TL+RS mixing strategy and evaluate models without any fine-tuning (2nd row in Table~\ref{tab:fewshot-result-transformer}/\ref{tab:fewshot-result-bart-simple}), we can observe a clear drop in the performance of randomly initialized model (Table~\ref{tab:fewshot-result-transformer}). 
We believe it is because using random spans for targets worsens the phraseness of the predictions. 
BART initialized models, on the other hand, show robust performance against these noisy targets.

\subsubsection{Performance in Low-Data Setting}
\label{sec:fewshot}

\begin{table*}[!ht]
    \renewcommand{\arraystretch}{0.4}
    \setlength{\tabcolsep}{0.3em}
    \small
    \centering
    \begin{tabular}{c|C{0.5cm}C{0.5cm}C{0.5cm}|C{1.5cm}|C{1.5cm}|C{1.5cm}|C{1.5cm}|C{1.5cm}}
    \toprule
        ~ & PT & DA & FT & \kpk & \openkp & \kptimes & \stackex & Avg \\ \midrule
        \multirow{2}{*}{0-shot} & x & ~ & ~ & 15.0 & 10.0 & 9.1 & 14.8 & 12.2 \\
        & x & x & ~ & 11.2 & 4.6 & 7.7 & 4.3 & 6.9 \\ \midrule
        
        \multirow{4}{*}{100-shot} & ~ & ~ & x & 0.5 & 0.2 & 2.4 & 5.1 & 2.1 \\
        ~ & ~ & x & x & 14.1 & 5.6 & 5.3 & 11.7 & 9.2 \\
        ~ & x & ~ & x & 14.5 & 20.1 & 22.6 & 13.0 & 17.6 \\
        ~ & x & x & x & 16.7 & 24.4 & 22.0 & 18.4 & 20.4 \\ \midrule
        
        \multirow{4}{*}{1k-shot} & ~ & ~ & x & 0.5 & 0.6 & 5.4 & 7.0 & 3.4 \\ 
        ~ & ~ & x & x & 15.0 & 8.6 & 8.9 & 15.4 & 12.0 \\ 
        ~ & x & ~ & x & 17.6 & 25.5 & 30.5 & 21.1 & 23.7 \\ 
        ~ & x & x & x & 19.7 & 28.0 & 30.7 & 26.3 & 26.2 \\ \midrule
        
        \multirow{4}{*}{10k-shot} & ~ & ~ & x & 3.4 & 1.5 & 19.2 & 20.8 & 11.3 \\ 
        ~ & ~ & x & x & 16.5 & 13.1 & 13.4 & 23.4 & 16.6 \\
        ~ & x & ~ & x & 20.6 & 30.6 & 38.6 & 31.4 & 30.3 \\
        ~ & x & x & x & 22.1 & 31.6 & 36.7 & 34.7 & 31.3 \\ \midrule
        \multirow{4}{*}{Avg} & ~ & ~ & x & 1.5 & 0.8 & 9.0 & 11.0 & 5.6 \\ 
        ~ & ~ & x & x & 15.2 & 9.1 & 9.2 & 16.8 & 12.6 \\ 
        ~ & x & ~ & x & 17.6 & 25.4 & \textbf{30.6} & 21.8 & 23.8 \\ 
        ~ & x & x & x & \textbf{19.5} & \textbf{28.0} & 29.8 & \textbf{26.5} & \textbf{25.9}\\ 
        \bottomrule
    \end{tabular}
    \vspace{-0.5em}
    \caption{Zero-shot and low-data results obtained by \tfrand. The best average score in each column is \textbf{boldfaced}.}
    \vspace{-1.0em}
    \label{tab:fewshot-result-transformer}
\end{table*}

\begin{table}[!h]
    \renewcommand{\arraystretch}{0.6}
    \setlength{\tabcolsep}{0.3em}
    \small
    \centering
    \begin{tabular}{c|C{0.5cm}C{0.5cm}C{0.5cm}|C{1.6cm}}
    \toprule
        ~ & PT & DA & FT & Avg \\ \midrule
        \multirow{2}{*}{0-shot} & x & ~ & ~ & 12.2 \\
        & x & x & ~ & 12.0 \\ \midrule
        \multirow{4}{*}{\makecell{Average of \\ few-shot \\ (100/1k/10k)}} & ~ & ~ & x & 36.2 \\
        & ~ & x & x & 36.3 \\
        & x & ~ & x & \textbf{36.6} \\ 
        & x & x & x & 36.1 \\ \bottomrule
    \end{tabular}
    \caption{Zero-shot and low-data results of \tfbart model. Full results are reported in appendix Table~\ref{tab:appendix:fewshot-result-bart}.}
    \label{tab:fewshot-result-bart-simple}
\end{table}

\begin{table}[!h]
    \centering
    \footnotesize
    \renewcommand{\arraystretch}{0.8}
    \begin{adjustbox}{width=0.48\textwidth}
    \begin{tabular}{C{0.8cm}|C{1.8cm}|C{1.2cm}|C{1.0cm}|C{1.2cm}}
    \toprule
        Model & DA Data & 100-shot & 1k-shot & 10k-shot \\ \midrule
        \multirow{3}{*}{\scriptsize \tfrand} & \kpk 100k & 16.7 & 19.7 & 22.1 \\ 
         & \magcs 1m & 16.8 & 19.4 & 21.8 \\ 
         & \magcs 12m & \textbf{17.6} & \textbf{20.4} & \textbf{22.8} \\ 
        \midrule
        \multirow{3}{*}{\scriptsize \tfbart} & \kpk 100k & 22.2 & 25.3 & 28.4 \\ 
        ~ & \magcs 1m & 22.3 & 25.4 & 28.4 \\ 
        ~ & \magcs 12m & \textbf{22.5} & \textbf{25.4} & \textbf{28.6} \\ 
    \bottomrule
    \end{tabular}
    \end{adjustbox}
    \caption{Average scores (over 4 datasets) with different amount of transfer labeled data for domain adaptation. All models are trained through three stages. The best score in each block is \textbf{boldfaced}.}
    \label{tab:scale-up-TL}
\end{table}
As described in \S\ref{sec:data_and_metric}, we use 100/1k/10k in-domain examples with gold standard keyphrases to fine-tune the model. 
To investigate the necessity of the \pt and \da stages given the \ft stage, we conduct a set of ablation experiments, skipping some of the training stages in the full pipeline. 


We start with discussing the results of randomly initialized models (Table~\ref{tab:fewshot-result-transformer}). 
\textbf{FT-only}: in the case where models are only fine-tuned with a small subset of annotated examples, models perform rather poorly on all datasets, especially on \kpk and \openkp, where more unique target phrases are involved. 
\textbf{DA+FT}: different from the previous setting, here all models are first trained with 100k pseudo labeled in-domain data points. We expect these pseudo labeled data to improve models on both phraseness and keyness dimensions. 
Indeed, we observe DA+FT leads to a large performance boost in almost all settings. 
This suggests the feasibility of leveraging unlabeled in-domain data using the proposed adaptation method (TL+RS). 
\textbf{PT+FT}: the pre-training stage provides a rather significant improvement in all settings, averaging over datasets and $k$-shot settings, PT+FT (23.8) nearly doubles the performance of DA+FT (12.6).
This observation indicates that the large-scale pre-training with domain-general phrase data can be beneficial in various down-stream domains, which is consistent with prior studies for text generation pre-training. 
\textbf{PT+DA+FT}: we observe a further performance boost when both PT and DA stages are applied before FT.
This to some extent verifies our design that PT and DA can guide the models to focus on different perspectives of KPG and thus work in an complementary manner. 

We also investigate when the model is initialized with a pre-trained large language model, i.e., BART~\cite{lewis2020bart}.
Due to space limit, we only report models' average scores (over the four datasets, and over the $k$-shot settings) in Table~\ref{tab:fewshot-result-bart-simple}, we refer readers to appendix Table~\ref{tab:appendix:fewshot-result-bart} for the full results.
We observe that in the pipeline, the fine-tuning stage provides \tfbart the most significant performance boost --- the average score is tripled, compared to the 0-shot settings, even performing solely the fine-tuning stage. 
This may be because the BART model was trained on a much wider range of domains of data (compared to Wikipedia, which is already domain-general), so it may have already contained knowledge in our four testing domains.
However, the auto-regressive pre-training of BART does not train particularly on the KPG task. This explains why it requires the BART model to fine-tune on KPG data to achieve higher performance.
The above assumption can also be support by further observations in Table~\ref{tab:fewshot-result-bart-simple}.
Results suggest that the DA stage is not notably helpful to \tfbart's scores, and the PT stage, on the other hand, seems to contribute to a better score.
We believe this is because the quality difference between labels used in these two stages: PT uses community-labeled phrases (high phrase quality but domain-general) and DA uses labels generated by the model itself (no guarantee on phrase quality but closer to target domains).
Since \tfbart only needs specific knowledge about the KPG task, the PT stage can therefore be more helpful.

We run Wilcoxon signed-rank tests on the results of Table~\ref{tab:fewshot-result-transformer}, and we find all differences between neighboring experiments (e.g. PT+FT vs. PT+DA+FT, both trained with KP20k and 10k-shot) are significant ($p<0.05$). For Table~\ref{tab:fewshot-result-bart-simple}, the improvement of PT+FT over the other three settings is also significant.

    
\subsubsection{Scaling the Domain Adaptation}
One advantage of self-labeling is the potential to leverage large scale unlabeled data in target domains.
We also investigate this idea and build a large domain adaptation dataset by pairing an unlabeled dataset with pseudo labels produced by a \pt model.
To this end, we resort to the MAG (Microsoft Academic Graph) dataset~\cite{sinha2015overview} and collect paper titles and abstracts from 12 million scientific papers in the domain of Computer Science, filtered by `field of study'. 
The resulting subset \magcs is supposed to be in a domain close to \kpk, yet it may contain noisy data points due to errors in the MAG's data construction process.
We follow the same experiment setting as reported in the above subsections, except that in the DA stage we either use 1 million or 12 million pseudo-labeled MAG data points for domain adaptation.
We train the models with the PT+DA+FT pipeline and report models' scores on \kpk test split.


As shown in Table~\ref{tab:scale-up-TL}, compared to our default setting which uses 100k unlabeled \kpk data points for domain adaptation, larger scale domain adaptation data can indeed benefit model performance --- models adapted with \magcs 12m documents show consistent improvements. 
However, the \magcs 1m data (still 10 times the size of \kpk) does not show clear evidence being helpful. 
We suspect the distribution gap between the domain adaptation data (i.e., \magcs) and the testing data (i.e., \kpk) may have caused the extra need of generalization.
Therefore, the \magcs 12m data may represent a data distribution that has more overlap with \kpk and thus being more helpful.
We also observe that models initialized with BART are more robust against such a distribution gap, on account of BART's pre-training with large scale of text in general domain.


\subsubsection{Multi-iteration Domain Adaptation}
\label{sec:multi-iter-da}

\begin{figure}[htb!]
    \centering
    \includegraphics[width=0.40\textwidth]{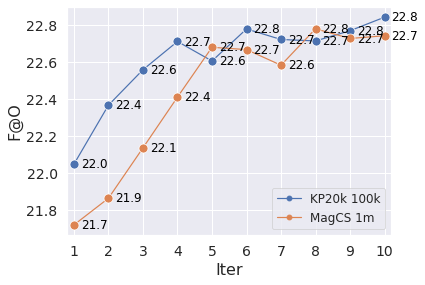}
    \vspace{-1.2ex}
    \caption{Trend of 10k-shot performance on \kpk with iterative self-labeling for 10 iterations.}
    \vspace{-0.5ex}
    \label{fig:multi-iter-self-training}
\end{figure}

Prior self-training studies have demonstrated the benefit of multi-iterations of label propagation~\cite{triguero2015selflabel,li2019learning2selftrain}. 
We conduct experiments to investigate its effects on KPG. 
Specifically, we first pre-train a \tfrand model using Wikipedia data as in previous subsections.
Then, we repeatedly perform the domain adaptation stage multiple times.
In each iteration, the model produces pseudo labels from the in-domain documents and then train itself with this data. 
Finally, we fine-tune the model with 10k annotated data points, and report its test scores on \kpk.
We consider two datasets, \kpk and \magcs 1m, as the in-domain data for domain adaptation.
As illustrated in Figure~\ref{fig:multi-iter-self-training}, the \tfrand model can gradually gain better test performance by iteratively performing domain adaptation using both datasets.
Due to limited computing resources, we set the maximum number of iterations to 10. But the trend suggests that models may benefit from more DA iterations.



\section{Related Work}

\paragraph{Keyphrase Generation.}
\citet{meng-etal-2017-deep} first propose KPG, which enables models to generate keyphrases according to importance and semantics, rather than extracting sub-strings from the text \cite{witten1999KEA,liu2011gap,wang2016ptr}.
Following this idea,
\citet{Chen2019TitleGE,wang2019topic,liang-etal-2021-unsupervised} propose to leverage extra structure information (e.g., title, topic) to guide the generation.
\citet{chan-etal-2019-neural,luo-etal-2021-keyphrase-generation} propose a model using reinforcement learning, and \citet{swaminathan2020preliminary} propose using GAN for KPG.
\citet{ye-etal-2021-one2set} propose to dynamically align target phrases to eliminate the influence of target phrase order, a problem highlighted by \citet{meng-etal-2021-empirical}.
\citet{mu2020keyphrase,liu2020reinforced,park-caragea-2020-scientific} use pre-trained language models for better representations of documents.
In a similar vein, \citeauthor{ye-wang-2018-semi} utilize self-learning to generate synthetic phrases for data augmentation, whereas we use self-labeling for domain adapation. \citeauthor{gao-etal-2022-retrieval} use a dense retriever to augment keyphrase generation in the cross-lingual scenario.

\paragraph{Pre-training for Phrase/Entity Understanding.} \citet{meng-etal-2021-empirical} show that pre-train models with noisy annotation can deliver great improvements on KPG. 
\citet{kulkarni2021learning} pre-train an understanding and a generation model with a large-scale annotated dataset OAGKX~\cite{ccano2020two} and the resulting models achieve decent performance on various NLP tasks. Both studies use a large amount of annotated data for pre-training, which is only available for certain domains. 
\citet{wang2021phrase,li2022uctopic} use contrastive learning to train phrase encoders. 
\citet{wang2021phrase,li2022uctopic} use contrastive learning to train phrase encoders. 
\citet{lee2021learning} find open-domain QA datasets can be used to 
 learn strong dense phrase representations.
Wikipedia is also frequently used in training models for entity-centric and knowledge-rich tasks. 
\citep{yamada2020luke,liu2021ner,xiong2019pretrained,meng2021distantNER,huang2021fewshotNER} use Wikipedia and its related resources as distant supervision to enhance BERT's abilities on modeling entities.

\paragraph{Self-labeling.} Self-labeling or self-training is a typical means for utilizing unannotated data and it has been applied in various machine learning tasks~\cite{he2019revisiting,mukherjee2020uncertainty}. 
\citet{yu2021fine} define rules as weak supervision for text classification and use self training to propagate labels to new documents. In our case, the pseudo labels are induced by models pre-trained with weak phrase annotation in Wikipedia. 
\citet{liang2020bond} use self-training to supplement distantly supervised NER and \citet{huang2021fewshotNER} use self-training to leverage unlabeled in-domain data.


\section{Conclusion}
In this study, we investigate domain gaps in the KPG task that hinder models from generalization. 
We attempt to alleviate this issue by proposing a three-stage pipeline to strategically enhance models' abilities on keyness and phraseness. 
Essentially, we consider phraseness as a domain-general property and can be acquired from Wikipedia data as distant supervision. 
Then we use self-labeling to distill the phraseness into data in a new domain, and the resulting pseudo labels are used for domain adaptation, as the labels can reflect the keyness and phraseness of the new domain. 
Finally, we fine-tune the model with limited amount of target domain data with true labels.
By taking the advantage of open-domain knowledge on the web, we believe this general-to-specific paradigm is generic and can be applied to a wide variety of machine learning tasks. 
As a next step, we plan to employ the proposed method for text classification and information retrieval, to see whether the domain-general phrase model can produce reliable class labels and queries for domain adaptation.

\section*{Limitations}
In this study, we provide empirical evidence of the impact of domain gap in keyphrase tasks, and we propose effective methods to alleviate it. However, we acknowledge that this study is limited in the following aspects: (1) As the first study discussing domain adapation and few-shot results, there is few studies to refer to as fair baselines. Nevertheless, we attempt to show the improvements of the proposed methods over base models by extensive experiments. (2) The pretrained keyphrase generation model can be used off-the-shelf, but  the multi-stage adaptation pipeline might increase the engineering complexity in practice. (3) We have only explored three strategies for domain adaptation, and they all require generating hard pseudo labels in different ways. Soft-labeling~\cite{liang2020bond} and knowledge distillation~\cite{zhou2021multi} methods are worth investigating. (4) We train a model with Wikipedia annotation to predict pseudo keyphrases, and it would be interesting to see if we can use large language models (e.g. GPT-3~\cite{gpt3}) to zero-shot predict phrases.

\section*{Ethics Statement}
\noindent \textbf{Dataset Biases} The domain-general pseudo phrases were produced based on public web-scale data (Wikipedia), and it mainly represents the culture of the Englishspeaking populace. Political or gender biases may also exist in the dataset, and models trained on these datasets may propagate these biases. Additionally, the pretrained BART models can carry biases from the data it was pretrained on.

\noindent \textbf{Environmental Cost} The experiments described in the paper primarily make use of V100 GPUs. We typically used four GPUs per experiment, and the first-stage pretraining may take up to four days. The backbone model \textsc{BART-large} 400 million parameters. While our work required extensive experiments, future work and applications can draw upon ourinsights and need not repeat these comparisons.

\bibliography{anthology,custom}
\bibliographystyle{acl_natbib}

\clearpage
\appendix

\section{Appendix}
\label{sec:appendix}

\subsection{Implementation Details}
\label{app:implementation}
Most experiments make use of four V100 GPUs. We elaborate the training hyper-parameters for reproducing our results in Table~\ref{tab:hp-tfrand} and \ref{tab:hp-tfbart}. For inference, we follow previous studies~\cite{yuan-etal-2020-one,meng-etal-2021-empirical} that uses beam search to produce multiple keyphrase predictions (beam width of 50, max length of 40 tokens).
We report testing scores with best checkpoints, which achieve best performance on valid set (2,000 data instances for all domains). 

\textit{Phrase masking ratio} denotes for \textit{p\%} of target phrases, replacing their appearances in the source text with a special token \texttt{[PRESENT]}.

\textit{Random span ratio} denotes replacing \textit{p\%} of words in the source text with a special token \texttt{[MASK]}.

\begin{figure}[!htb]
    \centering
    \includegraphics[width=0.4\textwidth]{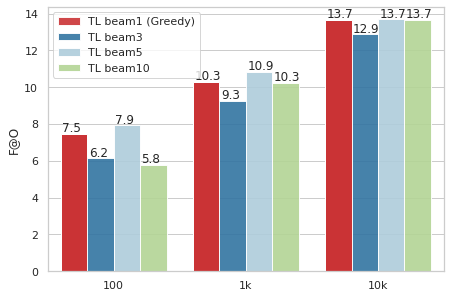}
    \caption{Comparison of domain adaptation with Transfer Labeling using different beam width to produce pseudo labels (\tfrand).}
    \label{fig:DA-beam-effect}
\end{figure}

\begin{figure}[!htb]
    \centering
    \includegraphics[width=0.4\textwidth]{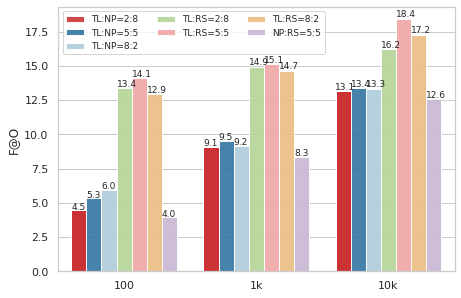}
    \caption{More results on mixing techniques for domain adaptation (\tfrand, with different mixing ratios). \textbf{TL}: Transfer Labeling. \textbf{NP}: Noun Phrases. \textbf{RS}: Random Span.}
    \label{fig:DA-mix-effect}
\end{figure}



\begin{table*}[h!]
  \small
  \centering
  \renewcommand{\arraystretch}{0.9}
  \begin{tabular}{l|ccccc}
    \toprule
    \bf Hparam & \bf PT & \bf DA & \textbf{PT+DA/PT+DA}$_{\magcs}$ & \bf FT 100/1k/10k & \bf *FT 100/1k/10k \\
    \midrule
    Max source length & 512 \\
    Max target length & 128 \\
    Max phrase number & 16 & 8 \\
    Max phrase length & 16 & 8 \\
    Phrase masking rate & 0.1 \\
    Random span ratio & 0.05 \\
    Batch size & $\approx$80 & 100 & 100 & 100 & 100 \\
    Learning rate & 3e-4 & 3e-4 & 1e-5 & 3e-4 & 1e-5  \\
    Number of steps & 200k & 40k & 20k/200k & 2k/4k/8k & 1k/2k/4k \\
    Warmup steps & 10\% \\
    Learning rate decay & linear \\
    Optimizer & Adam \\
    Adam $\beta_1$  & 0.9 \\
    Adam $\beta_2$  & 0.998 \\
    Adam epsilon  & 1e-6 \\
    Max gradient norm & 2.0 & 2.0 & 1.0 & 1.0 & 1.0 \\
    Dropout & 0.1 \\
    BPE Dropout & 0.1 & 0.0 & 0.0 & 0.0 & 0.0 \\
    Label smoothing & 0.1 \\
    Save checkpoint freq & ending step & ending step & ending step & 100/200/400 & 50/100/200 \\
    \bottomrule
  \end{tabular}
  \caption{\small
    Training hyperparameters for \tfrand. *FT denotes the fine-tuning stage in cases of PT+FT or PT+DA+FT. Empty cell means it is the same as the leftmost value.
  }
  \label{tab:hp-tfrand}
\end{table*}

\begin{table*}[h!]
  \small
  \centering
  \renewcommand{\arraystretch}{0.8}
  \begin{tabular}{l|ccccc}
    \toprule
    \bf Hparam & \bf PT & \bf DA & \textbf{PT+DA/PT+DA}$_{\magcs}$ & \bf FT 100/1k/10k \\
    \midrule
    Max source length & 512 \\
    Max target length & 256 & 256 & 256 & 128\\
    Max phrase number & 16 \\
    Max phrase length & 6 & 8 & 8 & 8 \\
    Phrase masking rate & 0.1 \\
    Random span ratio & 0.05 \\
    Batch size & 256 & 256 & 256 & 16 \\
    Learning rate & 1e-5  \\
    Number of steps & 40k & 5k & 5k/20k & 2k/4k/8k \\
    Warmup steps & 2.4k & 300 & 300/1.2k & 200/400/800 \\
    Learning rate decay & linear \\
    Optimizer & Adam \\
    Adam $\beta_1$  & 0.9 \\
    Adam $\beta_2$  & 0.98 & 0.98 & 0.98 & 0.999\\
    Adam epsilon  & 1e-8 \\
    Weight decay & 0.01 \\
    Max gradient norm & 1.0 & - & - & 0.1 \\
    Dropout & 0.1 \\
    Label smoothing & 0.1 \\
    Save checkpoint freq & ending step & ending step & 100/200/400 & 50/100/200 \\
    \bottomrule
  \end{tabular}
  \caption{\small
    Training hyperparameters for \tfbart. Empty cell means it is the same as the leftmost value.
  }
  \label{tab:hp-tfbart}
\end{table*}

\subsection{Data Statistics}
See Table~\ref{tab:appendix-dataset-statistics}.

\begin{table*}[!ht]
    \centering
    \small
    \begin{tabular}{l|r|r|r|r|c|c|c|c}
    \toprule
        ~ 
        & \makecell{\#doc}
        & \makecell{\#words \\in doc}
        & \makecell{\#kp}
        & \makecell{\#unique \\kp}
        & \makecell{\#kp \\per doc}
        & \makecell{\#uni kp \\per doc}
        & \makecell{\#present kp \\per doc}
        & \makecell{\#absent kp \\per doc}
        \\ \midrule
        \textbf{Training Sets} \\ \midrule
        Wikipedia & 3.2m & - & - & - & - & - & - & - \\
        \kpk & 514.2k & 161 & 2.7m & 680.1k & 5.3 & 1.3 & 3.3 & 1.9 \\
        \openkp & 134.9k & 1104 & 294.2k & 206.8k & 2.2 & 1.5 & 2.1 & 0.0 \\
        \kptimes & 259.9k & 803 & 1.3m & 104.8k & 5.0 & 0.4 & 2.4 & 2.6 \\
        \stackex & 299.0k & 207 & 803.9k & 8.1k & 2.7 & 0.0 & 1.6 & 1.1 \\
        \magcs 1M & 1.0m & 151 & 9.6m & 1.7m & 9.6 & 1.7 & 3.4 & 6.2 \\ 
        \magcs 12M$\dag$ & 12.1m & 151 & 115.9m & 14.3m & 9.6 & 1.2 & 3.4 & 6.2 \\ \midrule
        \textbf{Test Sets} \\ \midrule
        \kpk & 19,987 & 161 & 105.2k & 55.9k & 5.3 & 2.8 & 3.3 & 1.9 \\ 
        \openkp & 6,614 & 894 & 14.6k & 13.6k & 2.2 & 2.0 & 2.0 & 0.2 \\ 
        \kptimes & 10,000 & 804 & 50.4k & 13.9k & 5.0 & 1.4 & 2.4 & 2.6 \\ 
        \stackex & 16,000 & 205 & 43.1k & 4.5k & 2.7 & 0.3 & 1.6 & 1.1 \\ 
        \jptimes & 10,000 & 517 & 50.3k & 9.0k & 5.0 & 0.9 & 4.0 & 1.0 \\ 
        \duc & 308 & 701 & 2.5k & 1.8k & 8.1 & 6.0 & 7.9 & 0.2 \\ 
    \bottomrule
    \end{tabular}
    \caption{Statistics of training/testing datasets used in this study. $\dag$Only 7.7m papers in \magcs 12M have keyphrases.}
    \label{tab:appendix-dataset-statistics}
\end{table*}

\subsection{Additional Results and Analyses}
\label{app:more_results}
Figure~\ref{fig:DA-beam-effect} and \ref{fig:DA-mix-effect} show additional results of domain adaptation. In Figure~\ref{fig:DA-beam-effect}, we find that larger beam widths do not lead to significantly better scores after fine-tuning and thus we use simple greedy decoding for most of this study. 
In Figure~\ref{fig:DA-mix-effect}, we compare various domain adapation strategies of mixing different pseudo labels. Overall, we find that mixing labels of transfer labeling (\textbf{TL}) and random spans (\textbf{RS}) by 50\%:50\% leads to best performance.

\begin{figure*}[!b]
    \centering
    \includegraphics[width=0.8\textwidth]{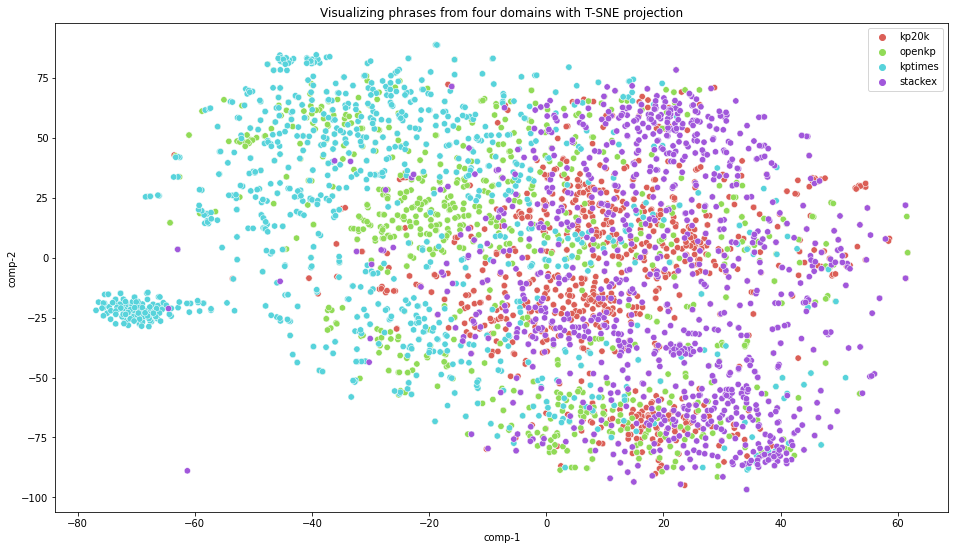}
    \caption{T-SNE visualization of keyphrase representations from four datasets.}
    \label{fig:appendix-visualize-domain-difference}
\end{figure*}

In Figure~\ref{fig:appendix-visualize-domain-difference}, we use T-SNE to visualize 1,000 most frequent keyphrases from each of four datasets (100k data examples from the training split) in the semantic space. We use BERT-base~\cite{devlin2019bert} to generate phrase embeddings (we feed forward each phrase independently as a sequence and take the \texttt{[CLS]} embedding as output). We use the T-SNE of Scikit-Learn~\cite{scikit-learn} with default hyperparameters. The result shows that phrases from each domain tend to gather into clusters. Particularly, we can see that a big overlap between \kpk and \stackex since both domains are related to Computer Science. The distribution of \openkp is more spread out, as its documents are collected from the web and cover a broader range of topics.

We present the full results of \tfrand and \tfbart in Table~\ref{tab:appendix:fewshot-result-transformer} and \ref{tab:appendix:fewshot-result-bart}. Besides, we supplement the evaluation with another two popular datasets: \jptimes (for models trained in the \jptimes domain) and \duc (for models trained in the \openkp domain).

\begin{table*}[!t]
    \renewcommand{\arraystretch}{0.5}
    \setlength{\tabcolsep}{0.3em}
    \small
    \centering
    \begin{tabular}{c|ccc|c|c|c|c|c|c|c}
    \toprule
        ~ & PT & DA & FT & \kpk & \openkp & \kptimes & \stackex & Average over 4 & \jptimes &  \duc \\\midrule
        \multirow{2}{*}{0-shot}
        & x & ~ & ~ & 15.0 & 10.0 & 9.1 & 14.8 & 12.2 & 15.8 & 9.4 \\ 
        & x & x & ~ & 11.2 & 4.6 & 7.7 & 4.3 & 6.9 & 12.7 & 6.6  \\ \midrule

        \multirow{4}{*}{100-shot}
        & ~ & ~ & x & 0.5 & 0.2 & 2.4 & 5.1 & 2.1 & 2.4 & 0.2 \\
        & ~ & x & x & 14.1 & 5.6 & 5.3 & 11.7 & 9.2 & 7.6 & 4.0  \\ 
        & x & ~ & x & 14.5 & 20.1 & 22.6 & 13.0 & 17.6 & 24.1 & 20.5  \\
        & x & x & x & 16.7 & 24.4 & 22.0 & 18.4 & 20.4 & 24.2 & 20.3 \\ \midrule
        \multirow{4}{*}{1k-shot}
        & ~ & ~ & x & 0.5 & 0.6 & 5.4 & 7.0 & 3.4 & 2.0 & 0.6 \\ 
        & ~ & x & x & 15.0 & 8.6 & 8.9 & 15.4 & 12.0 & 9.0 & 4.8 \\
        & x & ~ & x & 17.6 & 25.5 & 30.5 & 21.1 & 23.7 & 25.8 & 20.6 \\
        & x & x & x & 19.7 & 28.0 & 30.7 & 26.3 & 26.2 & 26.1 & 22.5 \\ \midrule
        \multirow{4}{*}{10k-shot}
        & ~ & ~ & x & 3.4 & 1.5 & 19.2 & 20.8 & 11.3 & 8.5 & 0.7 \\ 
        ~ & ~ & x & x & 16.5 & 13.1 & 13.4 & 23.4 & 16.6 & 9.6 & 6.4\\ 
        ~ & x & ~ & x & 20.6 & 30.6 & 38.6 & 31.4 & 30.3 & 25.7 & 24.3\\ 
        ~ & x & x & x & 22.1 & 31.6 & 36.7 & 34.7 & 31.3 & 27.1 & 23.6 \\ \midrule
        \multirow{4}{*}{Avg}
        & ~ & ~ & x & 1.5 & 0.8 & 9.0 & 11.0 & 5.6 & 4.3 & 0.5 \\ 
        ~ & ~ & x & x & 15.2 & 9.1 & 9.2 & 16.8 & 12.6 & 8.7 & 5.1 \\ 
        ~ & x & ~ & x & 17.6 & 25.4 & \textbf{30.6} & 21.8 & 23.8 & 25.2 & 21.8 \\
        ~ & x & x & x & \textbf{19.5} & \textbf{28.0} & 29.8 & \textbf{26.5} & \textbf{25.9} & \textbf{25.8} & \textbf{22.1} \\ 
        \bottomrule
    \end{tabular}
\vspace{-1ex}
\caption{Zero-shot and low-data results. Models are randomly initialized. The best average score is boldfaced.}
\vspace{-2ex}
\label{tab:appendix:fewshot-result-transformer}
\end{table*}

\begin{table*}[!h]
    \renewcommand{\arraystretch}{0.5}
    \setlength{\tabcolsep}{0.3em}
    \small
    \centering
    \begin{tabular}{c|ccc|c|c|c|c|c|c|c}
    \toprule
        ~ & PT & DA & FT & \kpk & \openkp & \kptimes & \stackex & Average over 4 & \jptimes &  \duc \\ \midrule
        
        \multirow{2}{*}{0-shot} & x & ~ & ~ & 14.7 & 9.7 & 10.5 & 13.9 & 12.2 & 16.3 & 9.8 \\ 
        ~ & x & x & ~ & 13.8 & 10.7 & 12.0 & 11.5 & 12.0 & 17.5 & 11.6 \\ \midrule
        \multirow{4}{*}{100-shot}
          & ~ & ~ & x & 22.3 & 32.8 & 31.6 & 29.6 & 29.1 & 27.9 & 16.6 \\ 
        ~ & ~ & x & x & 22.5 & 33.3 & 32.0 & 29.2 & 29.3 & 28.7 & 20.7 \\ 
        ~ & x & ~ & x & 22.4 & 33.7 & 31.6 & 31.1 & 29.7 & 28.4 & 21.5 \\ 
        ~ & x & x & x & 22.2 & 32.0 & 31.6 & 29.7 & 28.9 & 28.4 & 21.5 \\ \midrule
        \multirow{4}{*}{1k-shot}
          & ~ & ~ & x & 25.1 & 36.4 & 43.6 & 41.1 & 36.5 & 33.2 & 21.1 \\ 
        ~ & ~ & x & x & 25.3 & 36.2 & 43.2 & 40.9 & 36.4 & 31.8 & 21.0 \\ 
        ~ & x & ~ & x & 24.9 & 36.9 & 44.3 & 41.2 & 36.8 & 34.0 & 22.7 \\ 
        ~ & x & x & x & 25.3 & 36.5 & 42.9 & 40.1 & 36.2 & 31.9 & 22.1 \\ \midrule
        \multirow{4}{*}{10k-shot}
          & ~ & ~ & x & 28.2 & 40.8 & 53.3 & 49.3 & 42.9 & 34.4 & 23.2 \\ 
        ~ & ~ & x & x & 28.0 & 41.5 & 53.4 & 49.6 & 43.1 & 34.5 & 25.0 \\ 
        ~ & x & ~ & x & 28.2 & 41.3 & 53.4 & 49.7 & 43.1 & 34.2 & 25.0 \\ 
        ~ & x & x & x & 28.4 & 41.2 & 53.2 & 49.8 & 43.1 & 34.9 & 25.6 \\ \midrule
        \multirow{4}{*}{Avg}
          & ~ & ~ & x & 25.2 & 36.6 & 42.8 & 40.0 & 36.2 & 31.8 & 20.3 \\ 
        ~ & ~ & x & x & \textbf{25.3} & 37.0 & 42.9 & 39.9 & 36.3 & 31.7 & 22.2 \\ 
        ~ & x & ~ & x & 25.2 & \textbf{37.3} & \textbf{43.1} & \textbf{40.7} & \textbf{36.6} & \textbf{32.2} & 23.0 \\ 
        ~ & x & x & x & \textbf{25.3} & 36.6 & 42.6 & 39.9 & 36.1 & 31.8 & \textbf{23.1} \\ \bottomrule
    \end{tabular}
    \vspace{-1ex}
    \caption{Zero-shot and Few-shot results. Models are initialized with BART-large~\cite{lewis2020bart}. The best average score is boldfaced.}
    \vspace{-2ex}
    \label{tab:appendix:fewshot-result-bart}
\end{table*}

\end{document}